%% file: iclr2021_conference.tex
\documentclass{article} % For LaTeX2e
\usepackage{iclr2021_conference,times}

\input{math_commands.tex}

\usepackage{hyperref}
\usepackage{url}
\usepackage{xcolor}
\usepackage{graphicx}

\title{Feature Importance in a \\Deep Learning Climate Emulator}
\author{Wei Xu, Xihaier Luo, Yihui Ren, Ji Hwan Park \& Shinjae Yoo \\
Computational Science Initiative\\
Brookhaven National Laboratory\\
Upton, NY 11973, USA \\
\texttt{\{xuw,xluo,yren,parkj,sjyoo\}@bnl.gov} \\
\And
Balasubramanya T. Nadiga\\
Los Alamos National Laboratory \\
Los Alamos, NM 87545, USA \\
\texttt{balu@lanl.gov}
}

\iclrfinalcopy % Uncomment for camera-ready version, but NOT for submission.
\begin{document}

\maketitle
\lhead{AI: Modeling Oceans and Climate Change Workshop at ICLR 2021}

\begin{abstract}
  We present a study using a class of post-hoc local explanation methods i.e., feature importance methods for ``understanding'' a deep learning (DL) emulator of climate.
  Specifically, we consider a multiple-input-single-output emulator that uses a DenseNet encoder-decoder architecture and is trained to predict interannual variations of sea surface temperature (SST) at 1, 6, and 9 month lead times using the preceding 36 months of (appropriately filtered) SST data.
  First, feature importance methods are employed for individual predictions to spatio-temporally identify input features that are important for model prediction at chosen geographical regions and chosen prediction lead times.
  % These generated heatmaps not just highlight the impactful areas in the input images but also indicate their monthly contributions after aggregations.
  % Then after collecting the contribution heatmaps of the instances in the entire training set, we generate the average heatmaps as a global understanding of the model.
In a second step, we also examine the behavior of feature importance
in a generalized sense by considering an aggregation of the importance heatmaps over training samples.
  We find that: 1) the climate emulator's prediction at any
  geographical location depends dominantly on a small neighborhood
  around it; 2) the longer the prediction lead time, the further back
  the ``importance'' extends; and 3) to leading order, the temporal decay of ``importance'' is independent of geographical location.
  An ablation experiment is adopted to verify the findings.
  From the perspective of climate dynamics, these findings suggest a
  dominant role for local processes and a negligible role for remote
  teleconnections at the spatial and temporal scales we consider.
  From the perspective of network architecture, the spatio-temporal relations between the inputs and outputs we find suggest potential model refinements.
  %This in turn suggests that the thermal inertia of the ocean is the main source of predictability that permits the realized predictive skill. 
  We discuss further extensions of our methods, some of which we are considering in ongoing work.
\end{abstract}

\section{Introduction}
\label{intro}

\subsection{Climate Prediction with A Deep Learning Emulator}
Comprehensive climate models have emerged as a powerful tool in helping unravel and better understand the myriad  processes underlying climate and climate change. However, the immense computational costs associated with such comprehensive models preclude them from being used widely. As such, climate emulators that are built using data from simulations of such climate models (that are conducted at various national and international climate modeling centers) are of great interest and value.  It is in this context that we are interested in the problem of learning spatio-temporal variability of climate. While such learning can be achieved using both feedforward and recurrent networks \citep[e.g., see][]{nadiga2019predicting, parkmachine, jiang2019interannual, nadiga2021reservoir}, in this article, we restrict ourselves to considering a feedforward network that use convolutional layers.

If we assume that such a climate emulator has been built based on learning of spatio-temporal variability of climate (modeled or actual), a further necessity for its usage is that its behavior has to be robust. When, previously, statistical methods were used to build (linear) climate emulators, the inherent interpretability and parsimony of the statistical models ensured such robustness. In the realm of deep neural networks, however, such robustness cannot be assumed. This is notwithstanding the observation that  ``machine learning models tend to generalize well even though the number of parameters that have to be learnt may be far greater than the number of samples they have to be learnt from" across a wide range of areas of application \citep[e.g., see][]{zhang2017understanding,zhang2020identity}. As such, it seems prudent to conduct further tests of the machine learnt emulator to ensure such robustness.

On the one hand, in a broad sense, it is easier for a climate emulator (computational physics emulators in general) to be robust than it is, e.g., for a general purpose image classifier to be robust. This is because of the less diverse, high quality and controlled nature of the data on which the emulator is based. On the other hand, however, the chaotic, complex and multiscale nature of the dynamics of the physical system opens up other routes that can contribute to making the emulator less robust. While it is our intent to develop tools and tests for ensuring robustness of computational physics emulators, we presently report on the use of general purpose tools/tests that have been developed by the ML community at large when applied to a climate emulator that we have developed.

\subsection{Explainable Artificial Intelligence (XAI)}
XAI has emerged as an essential research direction in recent years to ``open the black box” of complex AI models and make them more understandable, trustworthy and controllable i.e., more robust. There are two major communities, AI and visual analytics (VA), trying to tackle the explainability and interpretability problem with their own preferences.

From the AI community, any non-inherently-interpretable model can be more transparent through post-hoc explanation. These explanation methods are grouped into local versus global explanations (\cite{aaaitutorial}). The local explanations seek for the understanding of individual predictions or in a local neighborhood of a given instance, while the global explanations aim at explaining overall behavior of the model, or engaging systematic-level biases affecting larger groups of data. 

From the VA community, there are a number of comprehensive surveys summarizing the state-of-the-art works depending on various categorization criteria. These methods typically present an interactive visualization system tailored to understand a specific category of models (e.g., the convolutional structure). \citet{yuan2020survey} partitioned the VA works based on a typical machine learning pipeline for real-world applications: data preprocessing before model building, machine learning model building, and deployment after the model is built. \citet{hohman2019} summarized VA works based on its role in DL research i.e. what, when and how to visualize deep learning models. 

\subsection{Our Work}
We leverage one class of post-hoc local explanation methods that we call \textit{feature importance methods} for understanding a DL emulator of climate SST prediction. Specifically, the multiple-input-single-output emulator, adopting a DenseNet encoder-decoder structure, takes preceding 36 months SST images as input to predict SST images at 1, 6 and 9 month lead times separately as output. 

In order to explain the model prediction, both instance and group data explanations are included. First, given an individual instance of data at a user-specified geographical pixel location, feature importance methods are adopted to generate 36 heatmaps highlighting contributions of the preceding 36 months' input features. These generated heatmaps present monthly impactful areas in the input that influence the model prediction result. Then after collecting the contribution heatmaps of the instances in the entire training set, we generate the mean heatmaps as an overall understanding of the model. We conclude that: 1) the climate emulator's prediction at any geographical location depends dominantly on a small neighborhood around it; 2) the longer the prediction lead time, the farther back the ``importance'' extends; and 3) the model's overall monthly contribution is independent of geographical locations. An ablation experiment is also conducted to verify our findings. These findings, after discussing with domain scientist, confirm validity of using emulators for SST prediction while also suggest potential improvement and extension directions as future works.

The rest of paper is structured as follows. We introduce our DenseNet model in Sec. \ref{network} and specific explanation methods in Sec. \ref{method}. Sec. \ref{result} presents both instance and group data explanation results with the methods. Finally, Sec. \ref{conclusion} summarizes the paper and discusses future works.

\section{DenseNet for Climate Prediction}
\label{network}

\textbf{The Problem Setup and Data}: We consider the variability of SST in the
North Atlantic over the last 800 years of the pre-industrial control
(piControl; a simulation in which external forcing is held fixed)
simulation of the Community Earth System Model
\citep[CESM2;][]{danabasoglu2020community} as part of the sixth phase
of the Coupled Model Intercomparison Project (CMIP6). CESM2 is a
global coupled ocean-atmosphere-land-land ice model and the piControl
simulation we consider uses the Community Atmosphere Model (CAM6) and
the Parallel Ocean Program (POP2), and at a nominal 1$^o$ horizontal
resolution in both the atmosphere and the ocean; the reader is
referred to \cite{danabasoglu2020community} for details.  This data is
publicly available from the CMIP archive at
https://esgf-node.llnl.gov/projects/cmip6 and its mirrors.
Since we are interested in predicting interannual variations\footnote{We note that while the diurnal cycle and the annual seasonal cycle constitute much larger variations, they are easily predicted. They are therefore not considered here.} of the SST
based on data from the CESM2 simulation described above, we
consider a twelve month moving-window average of the monthly SST
field in what follows.

\textbf{Architecture}: We trained a DenseNet as our baseline model (\citet{huang2017densely}). The model is designed to provide pixel-wise predictions of SST with different time leads. Specifically, the inputs to DenseNet are the time series data $\mathbf{x} = [x_{k-36}, x_{k-35}, \dots, x_{k-1}]$ and the output is a future state $x_{k+(i-1)}$. Thus, each sample contains $36$ input images and $1$ output image ($i=1$, $6$ and $9$ month lead times are studied in this work). The overall architecture takes the form of downsampling and upsampling. Specifically, a composite function contains three consecutive operations: batch normalization (BN), followed by a rectified linear unit (ReLU), and a non-unit stride $3 \times 3$ convolution (Conv) is applied to reduce the size of high-dimensional input data in downsampling. Another composite function of four consecutive operations: BN, ReLu, followed by a bicubic interpolation, and a $3 \times 3$ Conv is applied to recover the coarse spatial resolution in upsampling. After extensive hyperparameter and architecture search, we derive a baseline network with 20 composition layers. The detailed architecture and configuration is shown in Appendix.

% We trained a DenseNet as our baseline model (\citet{huang2017densely}). The model is designed to provide pixel-wise predictions of SST with different time leads (1, 6 and 9 month lead times are studied in this work). 

\begin{figure}[h]
\begin{centering}
\includegraphics[width=0.98\linewidth]{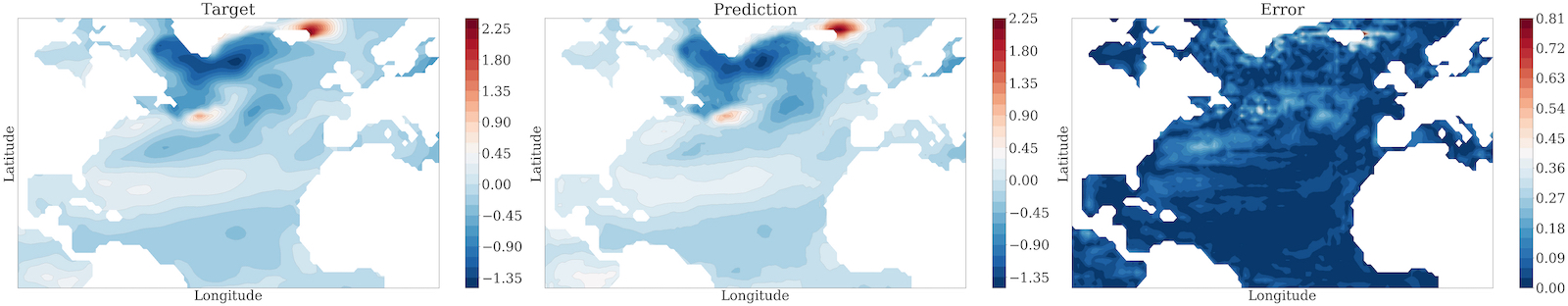}
\caption{Target, prediction, and error of a test sample for the DenseNet Climate model.}
\label{fig:testsample}
\end{centering}
\end{figure}

\textbf{Training}: 1280 samples were used to train the model, and another set of 2048 samples was used for validation. Specifically, mean squared error (MSE) was adopted as the objective function in the training process; L2 regularization or weight decay was considered to prevent overfitting, and Adam optimizer was utilized for optimization. Though the spatial dimension of a given input is $70 \times 125$, the predictive values locate in an irregular domain, i.e., only the oceans. Similar to many computer vision problems, we built a binary mask matrix for image segmentation. In our case, the mask matrix is applied to the prediction to remove non-valued locations that correspond to the land area. The mask matrix imposes hard constraints on model predictions during post-processing. The prediction of a test sample is shown in Fig. \ref{fig:testsample}.

\section{Post-hoc Local Explanation Methods}
\label{method}
\subsection{Feature Importance Methods}
The class of explanation methods that elucidates the internal model processes by highlighting relevant features in an input, typically an image, gains popularity recently due to its simplicity and insightfulness. These methods include gradient based approaches such as GradCAM (\cite{gradcam}) as well as decomposition approaches such as LRP (\cite{Bach2015}). Adebayo et al. named them Saliency methods (\cite{adebayo2020sanity}), while Lundberg et al. referred to feature attributions (\cite{lundberg2017unified}). In this work, we call this category of visualization and attribution methods \textit{feature importance methods}.

There have been debates over the superiority of different feature importance methods. For instance, although gradient based methods are easier to implement they suffer from the shattered gradients problem that decomposition approaches overcome but are less convenient to compute. Lundberg et al. (\cite{lundberg2017unified}) unified some popular decomposition approaches with two more renown model-agnostic methods, LIME (\cite{ribeiro2016why}) and Shapley values (\cite{trumbelj2013ExplainingPM}), as additive models and proposed new SHAP value estimation methods. 

In this work, we select a representative set of feature importance methods i.e., Guided Backprop, IntegratedGradients, DeepLIFT, DeepLiftShap that are better performed than others in our preliminary experiments: 

\textbf{Guided Backprop (GBP)} \quad GBP is similar to DeConvNet (\cite{deconvnet}) that it computes the gradient of target output with respect to the input, but negative gradients are set to zero when backpropagating through ReLU units (\cite{guidedbp}).

\textbf{IntegratedGradients (IG)} \quad IG is defined as $IG(x) = (x-x') \times \int_0^1 \frac{\partial f(x'+\alpha(x-x'))}{\partial x} \ d\alpha$, where $x$ is the input, $x'$ is a baseline input, $f$ is our model, and $\alpha$ is the scaling coefficient (\cite{integratedgrad}). It represents the integral of gradients with respect to inputs along the path from a given baseline to input. We choose zero baseline for this method.

\textbf{DeepLIFT (DLFT)} \quad  DLFT seeks to explain the difference in output from baseline in terms of the difference in input from baseline. It attributes to each input $x_i$ a value $C_{\Delta {x_i}\Delta o}$ that represents the effect of that input being set to a baseline value as opposed to its original value, where $\Delta {x_i}=x_i - {x'}_i$ is the input difference from baseline and $\Delta o = f(x) - f(x')$ is the output difference (\cite{lundberg2017unified}). DLFT uses the ``summation to delta" property $\sum_{i=1}^n C_{\Delta {x_i}\Delta o} = \Delta o$ meaning that the sum of all input changes equal to the output difference (\cite{deeplift}). We choose zero baseline for this method.

\textbf{DeepLiftShap (SHAP)} \quad DeepLiftShap is one of the few unified methods presented by Lundberg et al. to approximate Shapley values using DeepLIFT. It takes a distribution of baselines and computes the DeepLIFT attribution for each input-baseline pair and averages the resulting attributions per input example (\cite{captum}). We refer to this method as SHAP. We choose zero baselines.

\begin{figure}
\begin{centering}
\includegraphics[width=0.98\linewidth]{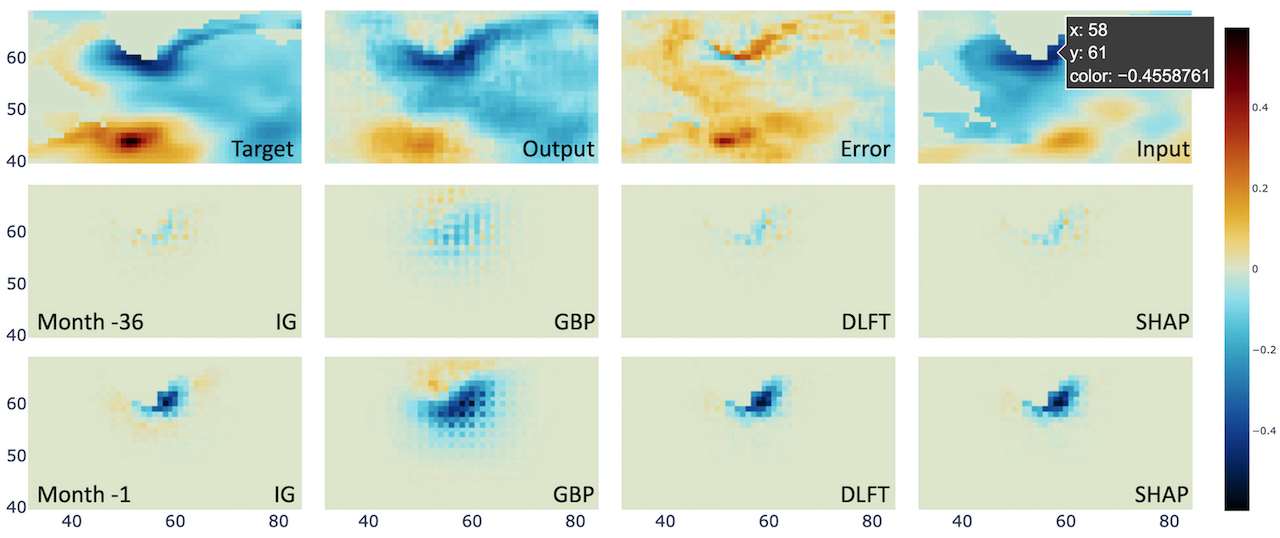}
\caption{Comparison among representative feature importance methods: IG, GBP, DLFT and SHAP. First row from left to right are the target, output, error and input images, where the input image is of month -36 with labeling the pixel location under study. Second and third rows show the heatmaps of these methods for month -36 and -1 respectively after zoom-in.}
\label{fig:all_methods}
\end{centering}
\end{figure}

\subsection{Adaption For Climate Model}

In order to apply these methods to explain our climate prediction model, necessary conversions are required. There are two major reasons behind. On one hand, the difference between our model and a general convolutional network that these methods are commonly applied to is that our model does not generate a single class score. This scalar score serves as the input to feature importance methods and is backpropagated along the pathway until the input layer and generates the output heatmap. However, our network generates a predicted image as output instead. One selected pixel of output image is interpreted each time as opposed to one class to interpret for a classifier. This modification makes our approach a pixel-wise explanation instead of a class-wise explanation.

On the other hand, compared to natural images, climate images include negative values. For feature importance methods, positive heatmaps reflect positive contributions from input features while negative heatmaps indicate negative contributions under a typical understanding. In our case, the negative values complicate the comprehension and could lead to contradictory conclusions. Therefore, we conducted preliminary experiments for all the combinations (positive/negative pixel values vs. positive/negative heatmap values) and learned: the input features highlighted by a positive heatmap are used to increase the output values while the ones by a negative heatmap are used to decrease the output values. Followup studies will be conducted to verify the observation.

In this work, we implement the aforementioned methods with Captum library as a built-in explanation tool for Pytorch models (\cite{captum}). Fig. \ref{fig:all_methods} compares the heatmaps for a location along the coast of Greenland. It is clear that IG and DLFT and SHAP are almost identical to each other and present precise maps consistent to the land contour while GBP tends to include a larger area due to shattered gradients. We choose DLFT in the rest experiments. 

%We compared these methods and picked a few. Show our result comparison. Eventually we picked deepshap or deeplift in the following experiment.

\section{Results}
\label{result}
We design several case studies to understand the model with both single instances and group of data. Single instance study explains local model behavior with a specific input and shows the prediction on individual geographical locations. Group data study takes the whole training dataset and explains the overall model behavior by aggregating the instance results.

\begin{figure}
\begin{centering}
\includegraphics[width=0.9\linewidth]{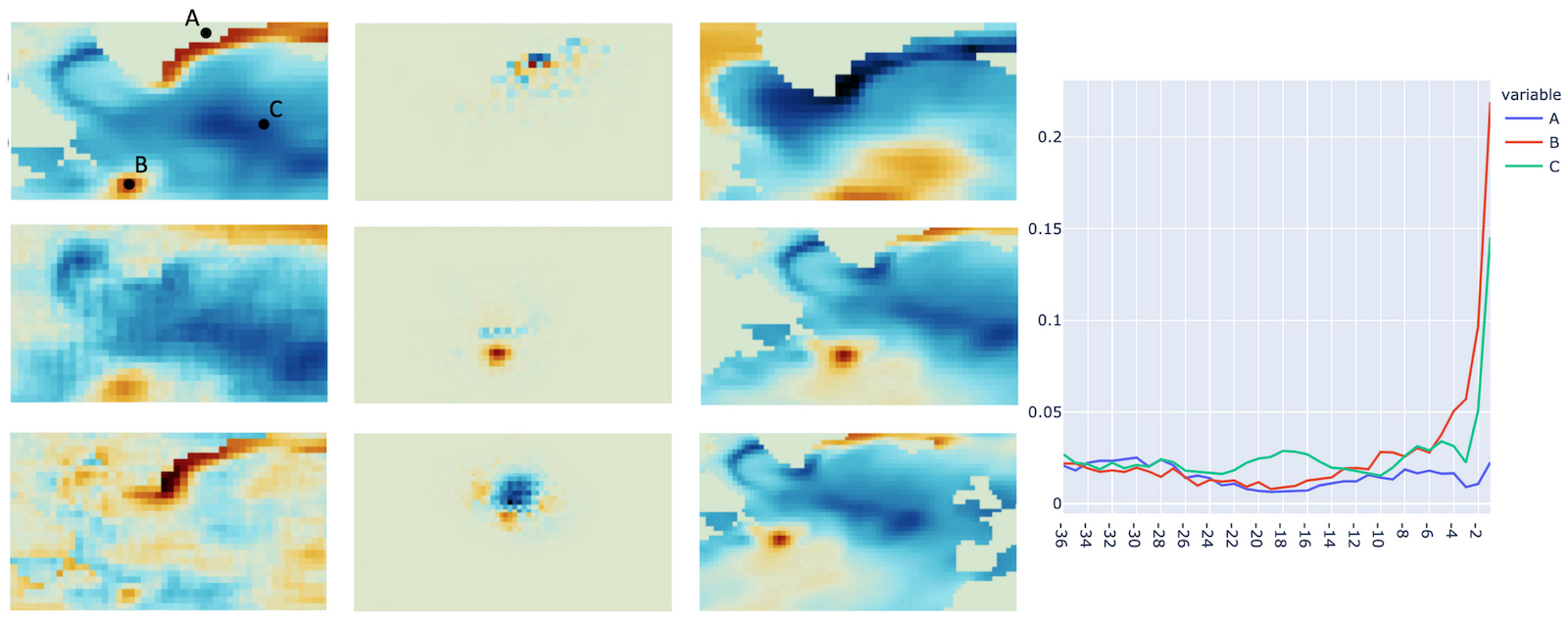}
\caption{Instance Explanation of the climate model. From top to bottom, the first column shows the zoom-in views for the target image with three selected locations labeled, the output image, and the error image; the second column shows the corresponding heatmaps for location A, B and C; the third column shows the corresponding input image for column 2; and the last column shows the monthly contributions for the three locations respectively.}
\label{fig:local_explain}
\end{centering}
\end{figure}

\subsection{Individual Instance Explanation}
We first focus on examining a single geographical location of a randomly selected input and consider two cases: 1) an ocean pixel vs. a land pixel, and 2) a positive pixel vs. a negative pixel. Therefore, three locations are selected as shown in the top left of Fig. \ref{fig:local_explain}, where $A$ is a land pixel, $B$ is a positive ocean pixel while $C$ is a negative ocean pixel. The model output of 1 month lead time is under study. The corresponding heatmaps generated for three locations are shown in the second column with their respective input images in column 3. Note that only one input month's heatmap is presented here. In the following, we discuss the strategy to select the specific months for different locations.

Since the input images are consecutive 36 months, the overall monthly contribution is essential to help us pick the most interesting months among them. Thus, we take the sum over the absolute values of a heatmap and obtain the accumulative contribution per month. Column 4 of Fig. \ref{fig:local_explain} shows the monthly contribution plot for three locations, where $x$ axis is the month index and $y$ axis is the accumulated contribution value. The most contributed month in each case (i.e., month -31, -1 and -1 respectively) was selected to show in column 2 and 3 of Fig. \ref{fig:local_explain}. Here we use negative month index to represent preceding months.

To sum up, by considering various locations and generating monthly heatmaps and monthly plots, we observed the following model behaviors:
\begin{itemize}
    \item The heatmaps suggest our network only focuses on a neighborhood of the target pixel for lead time prediction. Followup work is needed to verify this observation.
    \item As described in Sec. \ref{network}, the output for a land area is masked to a close zero value after the post-processing. The network itself still predicts non-zero values to the land as shown in the output image (Fig. \ref{fig:local_explain} column 1 middle) especially along the shore. The highlighted area (Fig. \ref{fig:local_explain} row 1 column 2) by the explanation method indicates where the value is from. 
    \item The monthly plot shows the prediction difference of land and ocean locations. Both ocean locations B and C take the closest month as most contributed one, while the land location A has no tendency to any month.
\end{itemize}

\subsection{Group Data Explanation}

\begin{figure}
\begin{centering}
\includegraphics[width=0.98\linewidth]{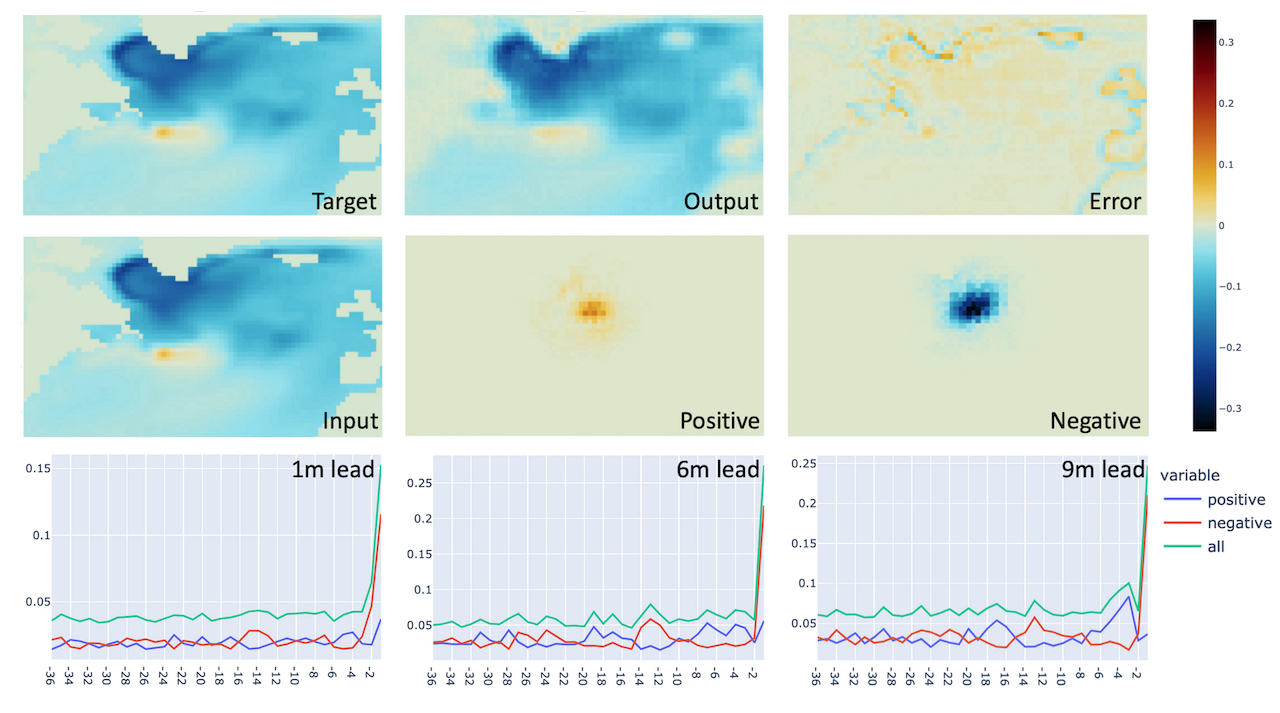}
\caption{Group data explanation over the training set to generate mean images of target, output, error, input (month -36 for illustration), positive and negative heatmaps. The monthly contribution plots are also shown for 1-, 6-, 9-month lead times.}
\label{fig:global_explain}
\end{centering}
\end{figure}

Then we study the group data (whole training set) to verify whether the observations of local explanations are random behaviors related to specific instances or the overall behavior of the model. Given a pixel location, heatmaps of all training inputs are collected. We split positive and negative heatmaps and accumulate them across the whole training set separately to generate mean positive and mean negative contribution heatmaps. The mean heatmaps are in a monthly fashion so that a mean monthly contribution plot can also be derived.

The following cases are considered: 1) the mean positive and negative monthly contribution of the training data; 2) for various ocean pixel locations, the comparison of mean contributions; 3) for various lead time predictions, the comparison of the mean contributions. The results are illustrated in Fig. \ref{fig:global_explain}. 

First, we pick one ocean pixel location around the upper middle area that is of high interest to our domain scientist to create mean positive and negative heatmaps for all 36 months respectively. Row 2 of Fig. \ref{fig:global_explain} (middle and right images) present the results for the most contributing month (-1 month) of 1-month lead time output. Just like single instance, the heatmaps highlight only the neighborhood around the selected pixel. Mean target, output, error and input over the entire training set are also visualized as reference. Second, the sum of absolute value of each heatmap (positive or negative) is computed to represent monthly accumulated contribution in order to create the mean monthly contribution plot. Both positive and negative monthly contribution plots are summed to create the total monthly contribution plot (referred as ``all" in the figures). The plots for three lead time outputs (1-, 6-, 9-month) results are shown in the last row of Fig. \ref{fig:global_explain}.

Finally, more user specified ocean pixel locations are chosen to replicate the same study. By comparing the contribution plots over various locations, it is confirmed that the patterns are almost identical for any lead time output. We plan to work on a more comprehensive study to compare all ocean pixels in future work. 

To sum up, by aggregating the instance explanation of the entire training set, we observed:
\begin{itemize}
    \item The mean heatmap still suggests the network only focuses on a neighborhood of the target pixel for lead time prediction, which is consistent with individual instance observation. A final experiment will be shown in the next section.
    \item The mean monthly contribution plots indicate that when lead time is longer more preceding months are leveraged. For instance, in the last row of Fig. \ref{fig:global_explain} from left to right, beside the salient contribution from month -1, the curves become more ``spiky" showing more and stronger impacts from other months. This observation meets the expectation of the domain scientist and confirms the necessity of including interannual data when lead time gets longer. 
    \item According to our preliminary study, the monthly contribution plot is independent of the location, together with the first observation, which may suggest a simplified model design.
\end{itemize}

\subsection{Ablation Experiment}
\label{ablation}

\begin{figure}
\begin{centering}
\includegraphics[width=0.65\linewidth]{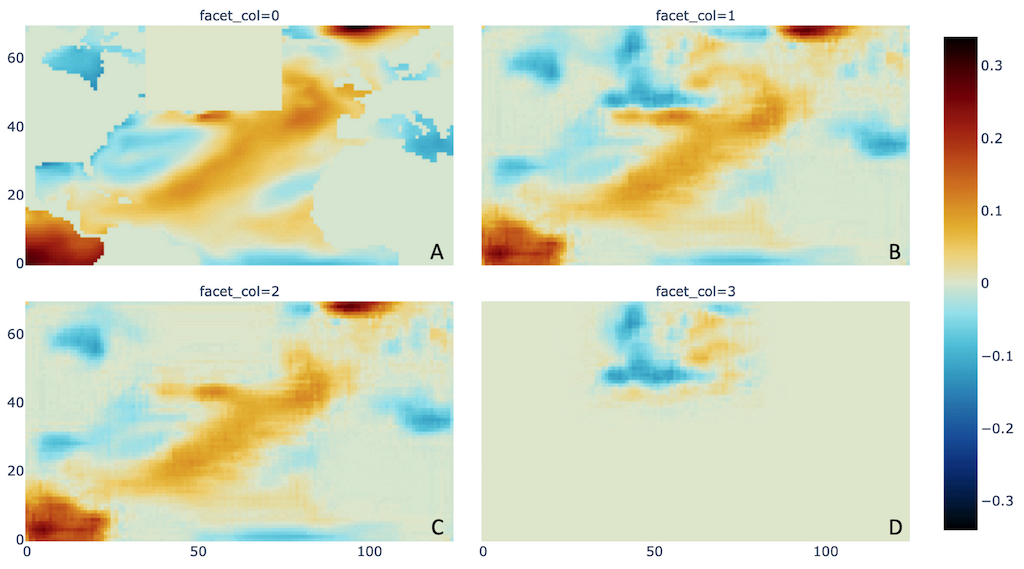}
\caption{Ablation experiment to test the contribution from non-neighbor regions: the input image of month -36 after ablating a small region (A), the output without ablation (B), the output after ablation (C), and the difference image (D).}
\label{fig:ablation}
\end{centering}
\end{figure}

To further verify the observation of previous case studies, we design an experiment to test the model output by ablating a small region in input images (Fig. \ref{fig:ablation}A). The output before (Fig. \ref{fig:ablation}B) and after ablation (Fig. \ref{fig:ablation}C) are subtracted to obtain the difference image (Fig. \ref{fig:ablation}D). We found that except a small surrounding area along the boundary of the ablated region the rest image has literally zero influence, while the surrounding area is affected due to the impact from nearby pixels. This experiment further confirms our observations in that this baseline model neglects teleconnections. Through discussing with domain scientist, this conclusion is contradictory with domain knowledge where the influence should be taken from long distance locations. After examining the network, future modification such as including a linear layer is considered to meet with domain knowledge.

\section{Conclusion and Future Works}
\label{conclusion}
In this paper, we presented an explanation approach employing feature importance heatmaps to understand the prediction behaviors of a deep learning climate emulator. The explanation is pixel-wise by attributing the input features to one output pixel location. Various case studies were designed to examine the model through individual instances and group data.
From the perspective of climate dynamics, the main finding of locality in both spatial and temporal domains in the relationship between the input fields and the predictions indicates a dominant role for local processes and a negligible role for remote teleconnections at the spatial and temporal scales we consider. 
Future work will keep studying group data behavior and leverage the findings to refine network architecture.

\subsubsection*{Acknowledgments}
This work is supported by the U.S. Department of Energy, Office of
Science, Advanced Scientific Computing Research under Award Number
DE-SC-0012704. BTN was supported by the DOE/SC SciDAC program under
project 'Non-hydrostatic dynamics with multi-moment characteristic
discontinuous Galerkin (NH-MMCDG) methods' and by LANL's LDRD program.
%Use unnumbered third level headings for the acknowledgments. All
%acknowledgments, including those to funding agencies, go at the end of the paper.

\bibliography{iclr2021_conference}
\bibliographystyle{iclr2021_conference}

\appendix
 \section{Appendix}
 We follow the guidelines provided \citet{huang2017densely} and provide the most promising configuration below:

 \begin{table}[h]
 \caption{Network architecture}
 \label{densenet}
 \begin{center}
 \begin{tabular}{llll}
 \multicolumn{1}{c}{\bf Layers}  & \multicolumn{1}{c}{\bf Resolution} &  \multicolumn{1}{c}{\bf Number f parameters} &  \multicolumn{1}{c}{\bf Configuration}
 \\ \hline \\
 Input         & $ 36  \times 70 \times 125 $  & NA      & NA    \\
 Convolution   & $ 144 \times 35 \times 63  $  & 129600  & $ k5s2p2 $    \\
 Dense Block   & $ 192 \times 35 \times 63  $  & 70080   & $ K16L3 $    \\
 Downsampling  & $ 96  \times 18 \times 32  $  & 101952  & $ k1s1p0 \,\, \& \,\, k3s2p1 $    \\
 Dense Block   & $ 192 \times 18 \times 32  $  & 119136  & $ K16L6 $    \\
 Upsampling    & $ 96  \times 36 \times 64  $  & 101952  & $ nearest \,\, \& \,\, k3s1p1 $    \\
 Dense Block   & $ 144 \times 36 \times 64  $  & 49056   & $ K16L3 $    \\
 Upsampling    & $ 144 \times 70 \times 125 $  & 34524   & $ nearest \,\, \& \,\, k3s1p1 $    \\
 Output        & $ 1   \times 70 \times 125 $  & NA      & NA    \\
 \end{tabular}
 \end{center}
 \end{table}

 where $k$ denotes the size of the convolving kernel, $s$ denotes the stride of the convolution, $p$ denotes the zero-padding added to both sides, $K$ denotes the  growth rate in Dense block, and $L$ denotes the number of layers.

\end{document}

%% file: math_commands.tex
\usepackage{amsmath,amsfonts,bm}

\def\eqref#1{equation~\ref{#1}}
\def\1{\bm{1}}

\DeclareMathAlphabet{\mathsfit}{\encodingdefault}{\sfdefault}{m}{sl}
\SetMathAlphabet{\mathsfit}{bold}{\encodingdefault}{\sfdefault}{bx}{n}